\documentclass{article}

\usepackage{PRIMEarxiv}
\usepackage{multirow}

\usepackage[utf8]{inputenc} 
\usepackage[T1]{fontenc}    
\usepackage{hyperref}       
\usepackage{url}            
\usepackage{booktabs}       
\usepackage{amsfonts}       
\usepackage{nicefrac}       
\usepackage{microtype}      
\usepackage{lipsum}
\usepackage{fancyhdr}       
\usepackage{graphicx}       
\graphicspath{{media/}}     

\title{A Multi-channel Training Method Boost the Performance 
}

\author{
  Yingdong Hu\\
  Zhejiang University \\
  Hangzhou\\
  \texttt{3190103847@zju.edu.cn} \\
}

\begin{document}
\maketitle

\begin{abstract}
Deep convolutional neural network has made huge revolution and shown its superior performance on computer vision tasks such as classification can segmentation. Recent years, researches devote much effort to scaling down size of network while maintaining its ability, to adapt to the limited memory on embedded systems like mobile phone. In this paper, we propose a multi-channel training procedure which can highly facilitate the performance and robust of the target network. The proposed procedure contains two sets of networks and two information pipelines which can work independently hinge on the computation ability of the embedded platform, while in the mean time, the classification accuracy is also admirably enhanced.
\end{abstract}


\section{INTRODUCTION}
Deep learning networks have exhibited state-of-the-art performance in various machining learning tasks recently. Especially in computer vision area, there has been a bunch of break through in image classification and object detection tasks. However, in most of the latest works, wider and deeper networks are preferred in building a top-preforming system, whose training and operating procedure are both time consuming and computing resource hungry. Whereas, many embedded systems like smart phones and face recognition door locks, are limited in memory and time, which means huge and deep network are probably no longer well suit for applications like these. 

There have been several talented works attempting to design a shallower network with a same state-of-the-art performance. Most recent to our work is \cite{GaoHuang2018MultiScaleDN}\cite{JiahuiYu2019SlimmableNN}. \cite{GaoHuang2018MultiScaleDN} has produced a fantastic network structure which can adjust its depth under different device environments. They announced that shallower formulations of the pretrained network will be used if the equipment is under low battery or short for memory. However, this method has created large redundancy in network structure, since it trains fully-connected layers separately for each shallower model, while these layers are actually the most space consuming sections and remain useless if shallower models are out of usage. \cite{JiahuiYu2019SlimmableNN} came up with a slimmable network which doesn't slim itself in depth but in width. Whereas, a crossing model batch normalization procedure, which has been advocated by [] that is useless for enhancing model's top performance, is still needed for the purpose of boosting the ability of slim network versions. That is to say, in a global perspective, extra memory will be taken even it won't improve the overall performance of the model.

Similar to  \cite{GaoHuang2018MultiScaleDN}, we have come up with a reducible network which can be transferred to shallower forms under memory or time limited circumstance. However, differently, we use a bottom-top structure rather than a top-bottom design proposed by \cite{GaoHuang2018MultiScaleDN}, which means most redundancy, the extra fully-connect part of the model can be reduced. A GAN-like training procedure has been presented in this work, which has been proved to be beneficial to improve the accuracy of the entire model as well as leading to a faster converge. Generally speaking, our main contributions are as follows:

\begin{itemize}
  \item [1)] 
  Put forward a bottom-top depth changeable structure which can reduce its depth to suit the memory and computing ability or current equipment.      
  \item [2)]
  Proposed a two-channel training procedure which help boost the performance of the entire network, in the mean time lead the training process a faster converge. A GAN-like lose function is used during this training period, which enable it to enhance the accuracy of the large ensemble of models simultaneously.
\end{itemize}

\section{RELATED WORKS}
\label{sec:headings}

\textbf{Model Scaling} has been a long developed topic for computer vision society, because with the vision technique maturing, more and more neural networks are required to be arranged in an edge device. Two main branches of model scaling has been produced recently, one is to replace the computational costly operations to other economical procedures\cite{XiangyuZhang2018ShuffleNetAE}\cite{AndrewHoward2017MobileNetsEC}\cite{KaiHan2020GhostNetMF}\cite{ForrestIandola2016SqueezeNetAA}\cite{MingxingTan2019EfficientNetRM}, the other is to slash the unimportant parameters in a pretrained cumbersome network\cite{AnkitGoyal2021NondeepN}\cite{HattieZhou2019DeconstructingLT}\cite{AdamGaier2019WeightAN}\cite{WeiWen2016LearningSS}. Different from the two methods mentioned above, there have been some attempts to establish a alterable structure for variable circumstance\cite{GaoHuang2018MultiScaleDN}\cite{JiahuiYu2019SlimmableNN}. These works have advanced a fantastic idea of providing different size of models at different resource-limited stage, and networks proposed by them can be straightforwardly used on the edge embedded devices like especially mobile phones, of which the computing ability varies according to the battery condition.

\textbf{Generative Adversarial Networks} was firstly proposed in 2015 by Ian J. Goodfellow\cite{goodfellow2014generative}, after which hundreds of GAN-like structures have been advocated to settle specific problems such as super resolution\cite{XintaoWang2018ESRGANES}, human face beautifying[]. However, the core idea of generative adversarial network remains the same, that is using the combination of two adversarial loss function in the training procedure, as shown in Equation (\ref{GAN Loss}), which forces the ability of both the generator and discriminator to enhance simultaneously.
\begin{equation}
\label{GAN Loss}
Loss=E_{x\sim P_{data}(x)}[log(D(x))]+E_{z\sim P_{z}(z)}[log(1-D(G(z)))]
\end{equation}
However, the combination of two inverse loss function also makes the generative adversarial network extremely difficult to train. On the contrary, if these two loss functions cooperate with each other rather than remain hostile, the training procedure will be rather smooth and easier to converge, and that is the basic training technique employed in our network. Details of this training procedure will be discussed in Section \ref{sec:Methods}.

\textbf{Knowledge Distillation} is another accessible method for network reduction\cite{AdrianaRomero2015FitNetsHF}\cite{GeoffreyEHinton2015DistillingTK}. \cite{GeoffreyEHinton2015DistillingTK} has produced a knowledge transfer algorithm by training the target lightweight network using soft labels generated by a specialist cumbersome network, which is referred as the teacher network. Various specialists are created to teach the student network for specific classification knowledge, such as classification for mushroom species and classification for gender. \cite{GeoffreyEHinton2015DistillingTK} asserts that knowledge distillation exhibits admirable ability and possibility because it shares knowledge, which means parameter in neural network, with an ensemble of models that can perform as well or even better than a cumbersome network. While in our model scaling technique, the shallower form of network also shares knowledge and parameters with the whole cumbersome network, which means our algorithm illustrated in Section \ref{sec:Methods} can be a new formulation of knowledge distillation.

\section{METHOD}
\label{sec:Methods}
In this section, we detail the proposed bottom-top depth changeable framework and the two-channel training procedure to train different remarkable classification networks and fully-convolutional networks. First, we bring up a toy experiment which can simply clarify our main innovations. Second, we illustrate the specific network structure and the proposed hints algorithm to guide network throughout training procedure. Finally, the design of our experiment is introduced and training details are discussed.

\subsection{TOY EXPERIMENT}
We first define a simple convolutional neural network, or ConvNet as it is commonly called, to do classification on Cifar-10\cite{AlexKrizhevsky2009LearningML}, which is a remarkable dataset known for its low pixels pictures. This ConvNet is so simple that it contains only four convolutional layers and two fully-connected layers, which means it is even shallower than LeNet, the first known ConvNet in the world. However, this ConvNet exhibits great performance on Cifar-10. Details about this network is demonstrated in Table \ref{tab:toy-exp}.

We first train the toy network on Cifar-10 from scratch straightforwardly, while changing the input image pixels from 32 to 128, doubled every time. We train the network for 50 epochs with a initial learning rate of 0.001 which drops to one-tenth every 10 epochs. Secondly a fully-convolutional network is added in front of the toy net to increase its depth. Same training procedure is conducted. Finally, we train the toy net along with the fully-convolutional network, using a loss function shown in Function (\ref{toy Loss}). Details of the fully-convolutional network is shown in Table \ref{tab:SRCNN}.
\begin{equation}
\label{toy Loss}
Loss= f_{loss}(ConvNet(x),labels)+f_{loss}((Conv_{heads}+ConvNet)(x),labels)
\end{equation}

\begin{table}
\caption{The convolutional head parts of the toy experiment. It's a fully-convolutional network which takes example by the SRCNN\cite{dong2015image}. This network varies it's channel depth and kernel size at different stage, and reconstruct the image in the final steps.}
\begin{center}
\label{tab:SRCNN}
\begin{tabular}{|c|}
\hline
 Convolutional head network \\
\hline\hline
input \\
\hline
conv 3-64, kernel size 5 \\
conv 64-12, kernel size 1 \\
\hline
\hline
conv 12-12, kernel size 3 \\
conv 12-12, kernel size 3 \\
conv 12-12, kernel size 3 \\
conv 12-12, kernel size 3 \\
\hline
conv 12-64, kernel size 1 \\
\hline
\hline
conv 64-3, kernel size 9 \\
\hline
\end{tabular}
\end{center}
\end{table}

\begin{table}
\caption{The structure of the very simple convolutional neural network. It contains four convolutional layers along with two fully-connected layers. The depth of Conv-layers is 32, 32, 64, 64 in order, and the number of neuron in FC-layers is 4096 and 512, which sums to a whole parameter number of 8 millions.}
\begin{center}
\label{tab:toy-exp}
\begin{tabular}{|c|}
\hline
 A Smaller Network \\
\hline\hline
input \\
\hline
conv 3-32 \\
conv 32-32 \\
\hline
maxpool\\
\hline
conv 32-64 \\
conv 64-64 \\
\hline
maxpool\\
\hline
FC 4096-512\\
\hline
FC 512-10\\
\hline
softmax\\
\hline
\end{tabular}
\end{center}
\end{table}

\begin{table}
 \caption{Result of the toy experiment. Toy ConvNet means our toy classification network, one pipeline means the training procedure after adding a fully-convolutional network, and two pipelines means training the network with the proposed loss function. What's more, shallow means test the shallower network without the fully-convolutional networks while deep means with them.}
  \centering
  \begin{tabular}{cccc}
    \toprule
    Network     & Pixel Size &Pipelines     & Accuracy \\
    \midrule
    \multirow{3}*{Toy ConvNet}  & 32  & \multirow{3}*{/}& 82.19\%    \\
                    & 64 &&78.57\% \\
                    & 128 &&72.42\% \\
    \multirow{3}*{One pipeline} & 32  & \multirow{3}*{/}&82.04\%     \\
                    & 64 &&80.60\% \\
                    & 128 &&77.17\% \\
    \multirow{6}*{Two Piplines} & \multirow{2}*{32} &shallow&82.57\% \\
                                                   &&deep&82.62\%\\
                                & \multirow{2}*{64} &shallow&80.00\% \\
                                                   &&deep&80.40\%\\
                                & \multirow{2}*{128} &shallow&73.41\% \\
                                                   &&deep&74.61\%\\
    \bottomrule
  \end{tabular}
  \label{tab:toy-results}
\end{table}

The result shown in Table \ref{tab:toy-exp} is fascinating because it has shown the last two-pipeline procedure leads the toy net to a better performance, although the performance along with fully-convolutional network actually drops, which can be illustrated by the parameter global migration caused by the combinative loss function.


\subsection{ALGORITHM}
Based on the toy experiment, we further illustrate how to apply this multi-pipeline training algorithm on model reduction in this section. Since that the pure classification network in the final stage achieves equal or even better performance than training it separately, and the deeper form with a fully-convolutional head takes a little more computations whereas bear superior result, we can therefore apply it on model scaling procedure, by employing the deeper form of network when there is enough memory or computing resource but the shallower one when resource is limited or under low battery condition.

The loss function should be a criterion that combines the performance of all pipelines, in order to enhance the ability of all models embedded in the whole network during the optimization stage. Since that we need to scale down the size of model rather than merely train a ensemble of fixable networks, preference should be considered in the training period, which means simple addition has to be replaced by weighted array. Based on this idea, we design the loss function shown in Equation (\ref{Our Loss}).

\begin{equation}
\label{Our Loss}
Loss=\sum_{i=1}^{batch}\sum_{j=1}^{pipelines}W_{j}\times f_{loss}(P_{j}(x),labels)
\end{equation}

A weight parameter is timed before the loss of every training pipeline, which is formally set to 1 in our toy experiment. And every pipeline is a independent classification procedure however shares the fully-connect layers and partly convolutional layers, and that automatically equals 2 in toy experiment.

\begin{figure}
  \centering
  \includegraphics[width=6cm]{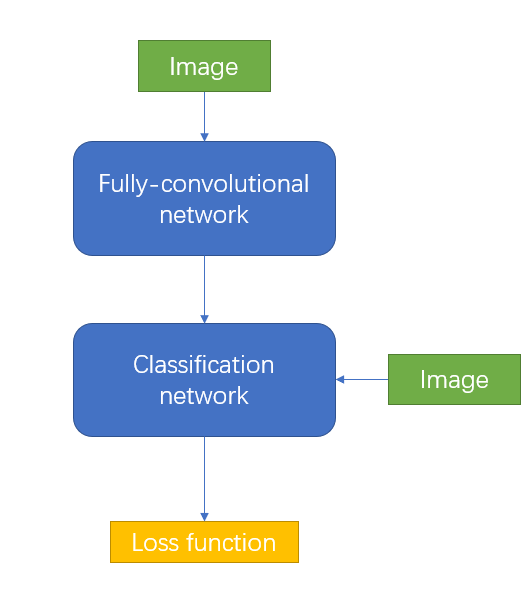}
  \caption{Sample figure caption.}
  \label{fig:fig1}
\end{figure}

Our framework resembles works in \cite{GaoHuang2018MultiScaleDN} as mentioned above. However, different from their work, which uses a huge periodic repeated fully-connected network as backbone and do classification every period using a specific fully-connected network, we do classification just once at the end, and reduce the fully-convolutional networks instead. That is to say, fully-connected network used by us share parameters from one training pipeline. The dissimilitude is clearly illustrated by Figure \ref{fig:fig1}.

\subsection{EXPERIMENT DESIGN}
We further verify our idea by carrying out series of experiments on some remarkable convolutional neural networks, especially the AlexNet, VGG and ResNet. Experiment are carried out on two famous dataset, the Cifar-10 and 
Cifar-100. Cifar-10 contains 10 classes and each of them has fifty thousands pictures, while Cifar-100 contains 100 classes with five thousands pictures of every class. Two fully-convolutional networks are applied to them independently. The structure of the shallower one is shown in Table \ref{tab:SRCNN}, name it the shallow convolutional network, while the deeper one is shown in Table \ref{VDSR}, name it the deep convolutional network.

Every training procedure is carried out on a Geforce RTX 3090 for 100 epochs, and the learning rate is initialized to 0.0001 with a cosine learning rate scheduler. For the experiment with two information pipelines, the network is trained together for only one round and test separately to get accuracy for models of different depth.

\begin{table}
\caption{A deeper form of convolutional head parts used in experiment. It's a fully-convolutional network which takes example by the VDSR\cite{kim2016accurate}. This network adds the residual images to the original image and reconstruct the image by the finally addition.}
\begin{center}
\label{VDSR}
\begin{tabular}{|c|c|}
\hline
\multicolumn{2}{|c|}{Deeper convolutional head network} \\
\hline\hline
\multicolumn{2}{|c|}{input} \\
\hline
residual&\multirow{9}*{original image}\\
\cline{1-1}
conv 3-128& \\
conv 128-128& \\
conv 128-128& \\
conv 128-128& \\
conv 128-128& \\
conv 128-128& \\
conv 128-128& \\
conv 128-3& \\
\hline
\multicolumn{2}{|c|}{add} \\
\hline
\end{tabular}
\end{center}
\end{table}

\section{RESULT}
The results shown in Table [\ref{tab:result0}][\ref{tab:result1}][\ref{tab:result2}] are done with the shallow convolutional head. Experiments for AlexNet goes without 32×32
image because there exists kernel size of 11 in it and information will vanish in the training pipeline.For the reason that the classification network has already performed well on the Cifar-10 network, the experiment done with deep convolutional head is only carried on ResNet18, Cifar-100, and the result is shown on Table \ref{tab:result3}.

\begin{table}
 \caption{AlexsNet result with the shallow convolutional head. / means this training procedure actually doesn't converge.}
  \centering
  \begin{tabular}{cccccc}
    \toprule
    Network     &Dataset & Pixel Size &Pipelines     & Accuracy &Baseline \\
    \midrule
    \multirow{8}*{AlexNet}&\multirow{4}*{Cifar-10}
                                & \multirow{2}*{64} &shallow&80.75\%&80.15\%  \\
                                                  &&&deep&80.88\%&78.36\%\\
                                 && \multirow{2}*{128} &shallow&77.47\%&86.63\%  \\
                                                  &&&deep&77.56\%&/\\
                             &\multirow{4}*{Cifar-100}
                                & \multirow{2}*{64} &shallow&52.50\%&51.32\%  \\
                                                   &&&deep&52.30\%&/\\
                                && \multirow{2}*{128} &shallow&64.81\%&59.41\%  \\
                                                   &&&deep&65.33\%&/\\
    \bottomrule
  \end{tabular}
  \label{tab:result0}
\end{table}

\begin{table}
 \caption{ResNet result with the shallow convolutional head.}
  \centering
  \begin{tabular}{cccccc}
    \toprule
    Network     &Dataset & Pixel Size &Pipelines     & Accuracy &Baseline \\
    \midrule
    \multirow{8}*{ResNet18}&\multirow{6}*{Cifar-100}  & \multirow{2}*{32} &shallow&51.06\%&51.04\%\\
                                                   &&&deep&51.15\%&47.66\%\\
                                && \multirow{2}*{64} &shallow&64.17\%&63.02\%  \\
                                                  &&&deep&64.35\%&59.33\%\\
                               && \multirow{2}*{128} &shallow&73.41\%&71.64\% \\
                                                   &&&deep&74.61\%& 69.37\%\\
                            &\multirow{2}*{Cifar-10}  & \multirow{2}*{128} &shallow&80.00\%&81.01\%  \\
                                                   &&&deep&80.03\%&78.90\%\\
    \bottomrule
  \end{tabular}
  \label{tab:result2}
\end{table}

\begin{table}
 \caption{VGG result with the shallow convolutional head. / means this training procedure actually doesn't converge.}
  \centering
  \begin{tabular}{cccccc}
    \toprule
    Network     &Dataset & Pixel Size &Pipelines     & Accuracy &Baseline \\
    \midrule
    \multirow{2}*{VGG}&\multirow{2}*{Cifar-100}  & \multirow{2}*{64} &shallow&76.1\%&76.8\%  \\
                                                   &&&deep&74.4\%&/\\
    \bottomrule
  \end{tabular}
  \label{tab:result1}
\end{table}

\begin{table}
 \caption{ResNet result with the deep convolutional head.}
  \centering
  \begin{tabular}{ccccc}
    \toprule
    Network     &Dataset & Pixel Size &Pipelines     & Accuracy \\
    \midrule
    \multirow{6}*{ResNet18}&\multirow{6}*{Cifar-100}  & \multirow{2}*{32} &shallow&51.61\%\\
                                                   &&&deep&52.04\%\\
                                && \multirow{2}*{64} &shallow&63.98\%  \\
                                                  &&&deep&63.82\%\\
                               && \multirow{2}*{128} &shallow&72.71\% \\
                                                   &&&deep&72.89\% \\
    \bottomrule
  \end{tabular}
  \label{tab:result3}
\end{table}

The result clearly explains that the boost of performance still exists on other remarkable networks and datasets, and the two-channel training procedure does improve the performance of the target network. According to this algorithm, a series of bottom-top reductive network structures could be established using a remarkable network as the backbone and a fully-convolutional network as a pre-process head which you can drop out under memory limited or low battery conditions.

This two-channel training procedure exhibits admirable possibility on image classification problem, and we own it to weights sharing, which is illustrated in \cite{GeoffreyEHinton2015DistillingTK} that weight sharing or the knowledge sharing can improve the performance. Compared to works in \cite{GaoHuang2018MultiScaleDN}, we share the whole fully-connected part of the network, where the most of parameters locates, which means there are more parameter and knowledge shared in our work.

\section{CONCLUSION}
We have proposed a new training procedure and fresh deductible network framework which can be used in model scaling area. A series of experiments are conducted to show that this observation is not a incident but can be transferred to various datasets and ConvNets. However, we only certify that this procedure only works well on classification problems, and how will it perform on segmentation or multi-object detection remains to be explore, but we have great confidence in that this procedure works equally well on other problems.

\bibliographystyle{unsrt}  
\bibliography{references}

\end{document}